\documentclass[10pt,final,journal,twocolumn]{IEEEtran}
\usepackage[utf8]{inputenc}
\usepackage{cite}
\usepackage[cmex10]{amsmath}
\usepackage{amsmath}
\usepackage{amsfonts}
\usepackage{amsthm}
\usepackage{algorithm}
\usepackage{algorithmic}
\usepackage{multirow}
\usepackage{multicol}
\usepackage{amsmath}
\usepackage{algorithm}
\usepackage{multicol,multienum}
\usepackage{multirow}
\usepackage[caption=false,font=footnotesize]{subfig}
\usepackage{color}
\usepackage{graphicx}
\usepackage{xspace}
\usepackage{tabularx}

\def\ie{\textit{i.e.}\xspace}

\def\etc{\textit{etc.}\xspace}
\def\eg{\textit{e.g.}\xspace}

\begin{document}

\title{Scarce Data Driven Deep Learning of Drones via Generalized Data Distribution Space}

\author{Chen~Li,~Schyler~C.~Sun,~Zhuangkun~Wei,~Antonios~Tsourdos~and~Weisi Guo
\thanks{Chen Li and Schyler C. Sun have equal contribution.

Chen Li, Schyler C. Sun, Zhuangkun Wei, Antonios Tsourdos and Weisi Guo are with Digital Aviation Research Technology Centre (DARTeC), Cranfield University, Bedford, United Kingdom. Weisi Guo is also with the Alan Turing Institute, London, United Kingdom. Funding: This work is supported by the Department of Transport under the S-TRIG program 2020-21; and the EPSRC/UKRI Trustworthy Autonomous Systems Node in Security [grant number EP/V026763/1]. Corresponding author: weisi.guo@cranfield.ac.uk.
}}

\maketitle

\begin{abstract}
Increased drone proliferation in civilian and professional settings has created new threat vectors for airports and national infrastructures. The economic damage for a single major airport from drone incursions is estimated to be millions per day. Due to the lack of diverse drone training data, accurate training of deep learning detection algorithms under scarce data is an open challenge. Existing methods largely rely on collecting diverse and comprehensive experimental drone footage data, artificially induced data augmentation, transfer and meta-learning, as well as physics-informed learning. However, these methods cannot guarantee capturing diverse drone designs and fully understanding the deep feature space of drones. Here, we show how understanding the general distribution of the drone data via a Generative Adversarial Network (GAN) and explaining the missing features using Topological Data Analysis (TDA) - can allow us to acquire missing data to achieve rapid and more accurate learning. 
We demonstrate our results on a drone image dataset, which contains both real drone images as well as simulated images from computer-aided design. When compared to random data collection (usual practice - discriminator accuracy of 94.67\% after 200 epochs), our proposed GAN-TDA informed data collection method offers a significant 4\% improvement (99.42\% after 200 epochs). We believe that this approach of exploiting general data distribution knowledge form neural networks can be applied to a wide range of scarce data open challenges.
\end{abstract}


\IEEEpeerreviewmaketitle

\section{Introduction}
\begin{figure*}
     \centering
     \includegraphics[width=1.0\linewidth]{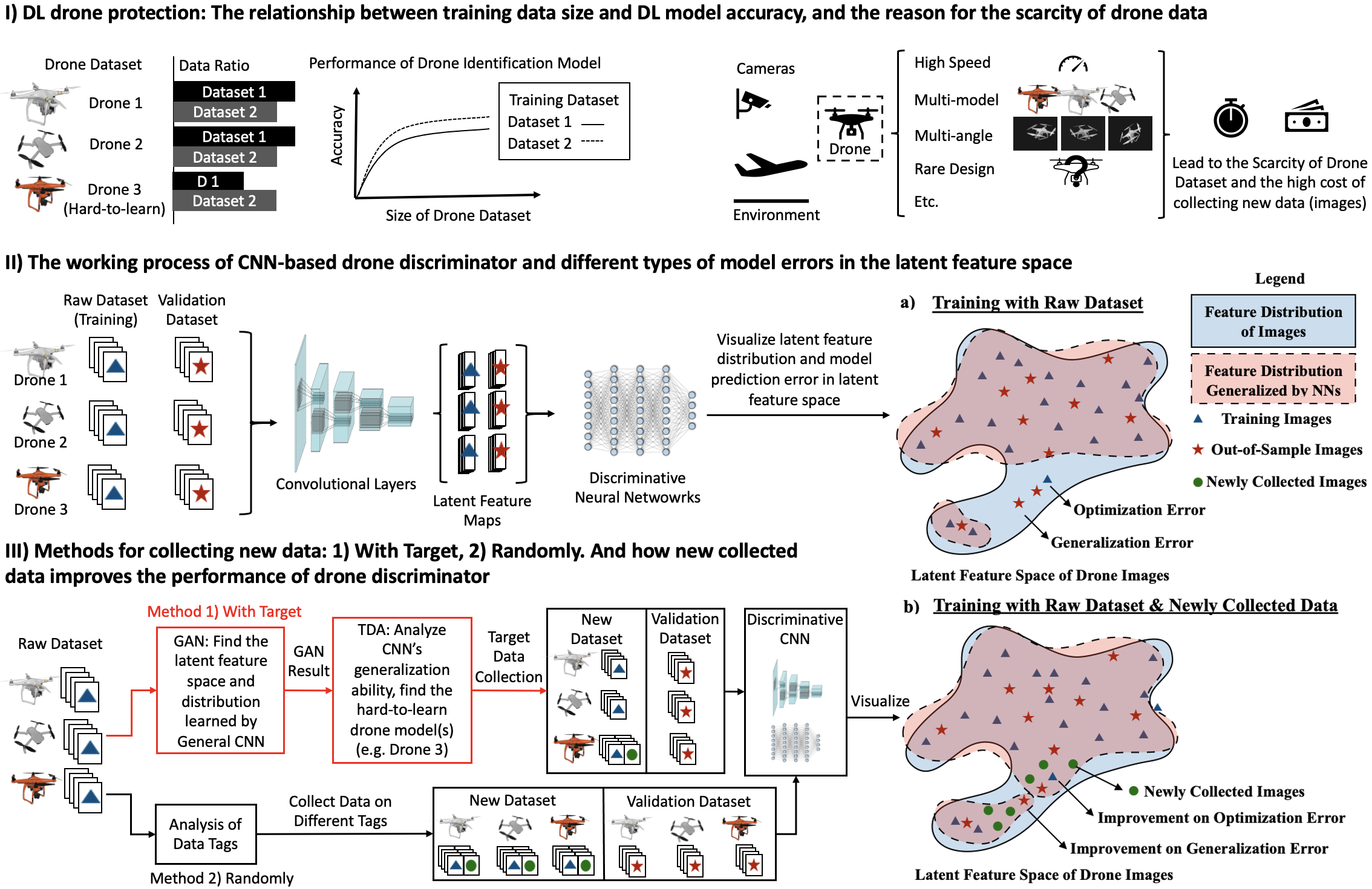}
    \caption{(I) The upper limit of DL drone classification accuracy is affected by the training data size, but various factors of drones lead to scarcity of drone data and high acquisition costs. (II) the nature of discriminative NNs’ inference is to map an
high-dimensional input features into a label class using the feature distribution (FD) generalized from the training data (convolutional layers extract latent features maps of input images for further discriminative work in NN). While lack of training data will result in the generalization error on out-of-sample data. (III) Our aim is to detect the data that important for model training (e.g. hard-to-learn drone), reduce the amount of new data required for model improvement, and outperform randomly data collection method in both learning speed and predication accuracy of trained model.}
     \label{intro}
\end{figure*}

\IEEEPARstart{I}{ncreased} proliferation of drones and autonomous air vehicles can disrupt critical national services ($\eg$ Gatwick Airport 2018). The economic damage for air transport is estimated to be millions per day for airports and airlines. Furthermore, the drone industry makes up 1.9\% of UK GDP and supports over 600,000 jobs in the drone’s economy. Whilst many are amateur drones that pose no malicious intention, some may carry deadly capability and cause severe economic damage to critical infrastructure. Protection against drones is critical to ensuring smooth operation of services, whilst safeguarding it against the most severe threats. High resolution cameras can classify drones using deep learning, but accurate identification is critical for not disrupting normal day-to-day operations and maintaining an efficient economy  \cite{PartC}. 
Whilst the upper limit of accuracy for image classification has been increased by more complex deep learning (DL) architectures, the upper accuracy of DL model also limited by its logarithmic growth to the size of the training dataset (see - Fig.\ref{intro}.\textit{I}) \cite{JFT-300M}. Simultaneously, lack of hard-to-learn (by DL model) drone data will affect the overall drone detection accuracy.
It is a key open challenge to achieve extremely high accuracy by sourcing sufficient, relevant but rare training data sets ($\eg$ rare drone design) \cite{Collection}. This often means a large amount of resources and time is dedicated to broad data collection and (re)training the neural network. Especially, high speed and tini-size of drone challenge the image capture, at the same time, multi-model and numbers of shooting angle increase the data scarcity of drone dataset - see Fig.\ref{intro}.\textit{I}. Therefore, the high cost of collecting new drone images adds to the cost of improving the accuracy of DL-based drone classification. However, a target of data collection can reduce the amount of new data to be collected compared to collecting data on all drones, saving overall DL performance improvement costs.


In this paper, we show how understanding the general distribution of the drone data via a Generative Adversarial Network (GAN) and explaining the missing features using Topological Data Analysis (TDA) - can allow us to acquire missing data to achieve rapid and more accurate learning. The aim is to demonstrate how the improved accuracy can benefit air transport protection.

\subsection{Related work}
One of the universal challenges in deep neural network training is when there is a lack of data, the out-of-sample performance can not be guaranteed \cite{FSLreview}. As shown in Fig.\ref{intro}.\textit{II}, the nature of discriminative NNs' inference is to map an high-dimensional input features into a label class using the feature distribution (FD) generalized from the training data \cite{ImageNet} (convolutional layers extract latent features maps of input images for further discriminative work in NN). While lack of training data will result in the generalization error on out-of-sample data \cite{GeneralizationError}, data scarcity can also lead to the optimization error on known training data (see - Fig.\ref{intro}.\textit{a}). However, collect extra training data could address these error as shown in Fig.\ref{intro}.\textit{b}. In general discriminative NNs, the performance increases logarithmically based on volume of training data size \cite{JFT-300M}, which means the marginal cost of training data for model performance improvement boosts exponentially. Due to the diminishing returns, exhaustive or randomly searching for data is not applicable to the scenarios where data collection is expensive ($\eg$ aerospace, military). Therefore, there is a need to create a method to identify which specific new data would be important for discriminative NNs' performance improvement based on an existing dataset, as the guidance to the new data collection work, so that to reduce data collection cost. This is the motivation for this paper. As shown in Fig.\ref{intro}.\textit{III}, our aim is to detect the data that important for model training (e.g. hard-to-learn drone), reduce the amount of new data required for model improvement, and outperform randomly data collection method in both learning speed and predication accuracy of trained model. There are related papers addressing the aforementioned challenges in the discriminative NNs training with limited data. We summarize several research achievements below and compare them in table \ref{comparison}.

Data augmentation improve the training set by adding slightly modified ($\eg$ translations, rotations and flips) copies of existing data to strengthen the invariance of NNs to the aforementioned modified data \cite{DataAugmentation}. However, data augmentation is only based on raw training data, hence cannot offer additional generalization on the variation of the object itself other than its position to NNs. Transfer learning reduce the training cost of new DL model by reusing the convolutional kernels from other related well-trained DL models \cite{TransferLearning}. By doing so, NNs can partially transfer the generalization ability got in the former relevant training into the latter target inference. Similarly, meta-learning tries to abstract more universal generalization ability from multiple training domains and “learn to learn” fast \cite{MetaLearning}, which lays the foundation for the few-shot learning \cite{Meta2FewShot}. However, the performance gain via these two methods are not guaranteed, since the transferred generalization ability is context-agnostic ($\ie$ does not focus on the properties of the new data), which may not match the need to specific requirements. Physical informed learning is designed to embed given laws of physics ($\eg$ general nonlinear partial differential equations) into NNs to inform its generalization \cite{PINN1,PINN2}. Hence, there is less need for the diversity of the training data and more training data can be generated numerically from the given laws of physics. However, in most discriminative NNs applications, physical law is unknown.

\begin{table*}[ht]

\centering

\begin{tabular}{|>{\centering}p{50pt}|>{\centering}p{75pt}>{\centering}p{75pt}>{\centering}p{75pt}>{\centering}p{75pt}>{\centering}p{75pt}|}
		\hline
         & {Data Augmentation \cite{DataAugmentation}} & {Transfer Learning \cite{TransferLearning}} & {Meta Learning \cite{MetaLearning}} & {Physical Informed Learning\cite{PINN1}} & {\textbf{Proposed GAN-TDA Method}}\cr \hline 
        Methodology & Add modified data into dataset & Reuse Layers from other well-trained DL models & Reuse layers trained in previous tasks & Apply physical properties in model training & Purposefully collect new data \cr \hline 
        Generalization Ability& Modified dataset & Other related data & Other related data & Data physical properties & Newly collected data \cr \hline 
        Advantage & No additional data required & Wide applicability; fast deploy speed & Learn to learn; learn fast for new tasks &  Controllable generalization & Explainability; less data to be collected; \cr \hline 
        Disadvantage & Limited improvement; lack of explainability & Context-agnostic, lack of explainability & Context-agnostic, lack of explainability & Difficulties in processing physical properties & Need additional cost for collecting targeted new data \cr \hline

\end{tabular}
\caption{A comparison of methods for training deep learning models with scarce data}
\label{comparison}
\end{table*}

Although data augmentation enhances the training set by adding augmented data based on observed data (auxiliary variables method), there is no effect on un-observed data (e.g. Drone images by un-observed angle). However, newly collected training set could address this problem. Transfer learning and meta-learning can save the training time for new tasks, but can not enhance a trained model and are lack of explainability. Physical informed learning needs expert experience which is abstract and uncontrollable. Thus, we propose a method to reveal the relationship between NN’s generalization ability and the composition of training data, so that to provide a feature-based target for new data collection to save the cost of collecting new data.



\subsection{Innovation: GAN-TDA framework}
In conventional work, methods focus on the generalization ability of the model itself, which are general methods with versatility for various applications. 
By contrast, the generalization ability brought by the training data attracts less attention. 
In this paper, we aim to identify the required data for discriminative NNs' generalization error reduction, by analyzing the existing training data through its potential feature distribution (FD) and that generalized by NNs (see - Fig.\ref{intro}.\textit{b}).

\paragraph{Generative adversarial network}
Generative models are designed to generate data with the same FD as that learned by NNs. In principle, most common generative models, including variational auto-encoder (VAE) and variations of generative adversarial network (GAN), are trained to convert the initial distribution of latent variables into that learned by NNs with training data \cite{VAE,VAE2020,GAN,WGAN,StyleGAN}.
For scarce data problems such as our drone detection application, one might be interested in using GAN with the following reasons: 
(1) GAN is proven to be asymptotically consistent in FD approximation while VAE may have bias due to the variational lower bound, thus the generated data from GAN would be more precise in representing the FD learned by the NNs \cite{Generator}; 
(2) With the same training data, the discriminator in GAN gives similar but more smooth convolution kernels as that in CNN \cite{Discriminator2}. More specifically, kernels in GAN have the same generalization way but weaker ability as CNN. Accordingly, the analysis in GAN can be considered as the representation of general CNN.
(3) The discriminator in GAN can be considered as the pre-trained target discriminative NN for methodology validation so as to eliminate generalization ability bias caused by another arbitrary NNs \cite{Discriminator1}.

In our experiment, color information in each pixel of drone images are use as inputs, the viewing of raw feature space built by pixel information will be over-dimensioned and lacks practical interpretability. Thus, to build a more informative feature space for image data, latent feature learned and processed by convolutional layers in deep learning models helps. Here, each latent feature is represented as the degree of response to a certain kernel feature in different receptive fields of the image. However, GAN does not give explicit expressions of the distribution, hence Monte Carlo synthetic data from the generator would be taken for the further FD analysis in the next step. 

\begin{figure*}[h]
     \centering
     \includegraphics[width=1.0\linewidth]{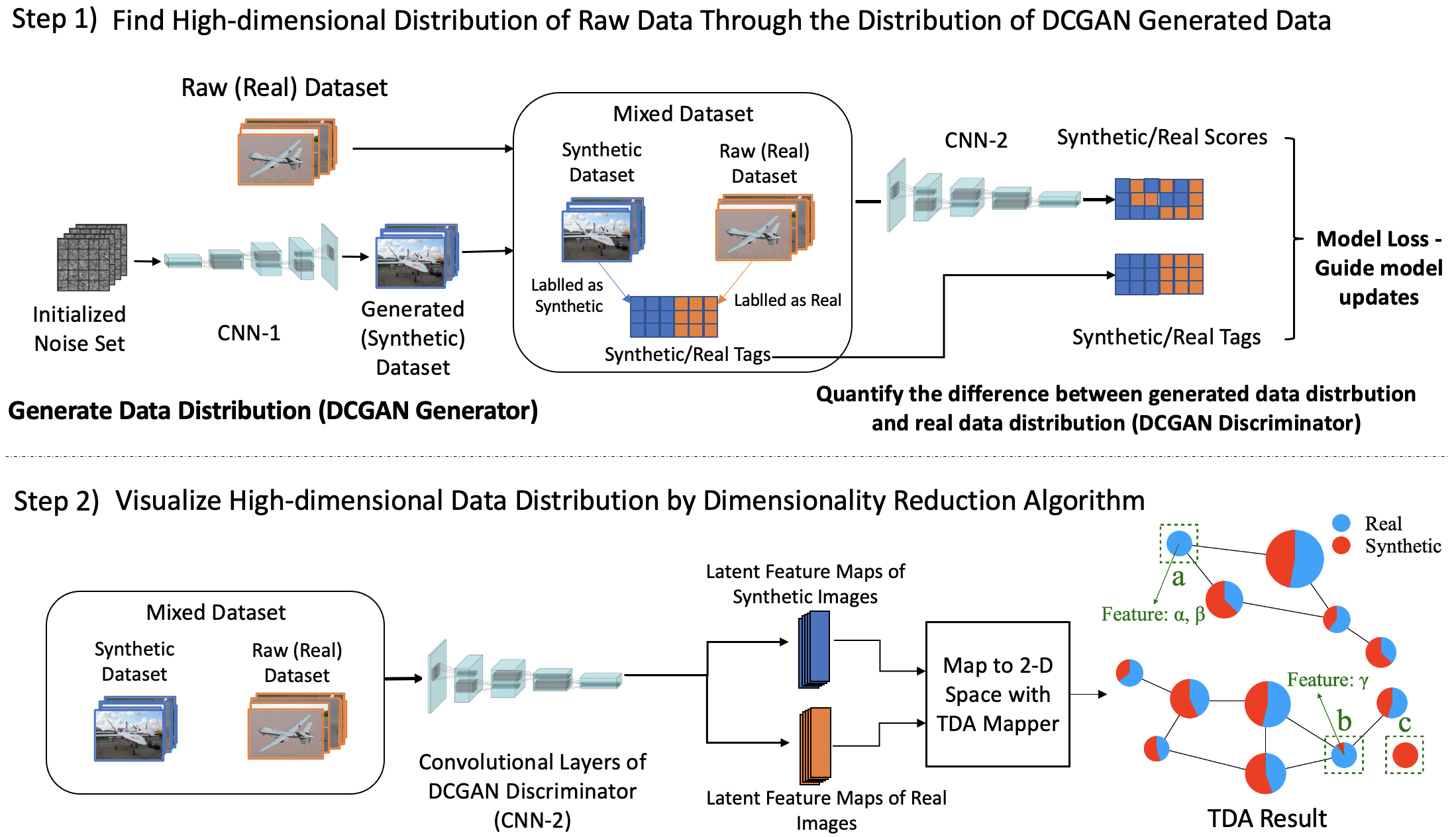}
     \caption{Demonstration of Method: Step 1) Find high-dimensional (latent features) distribution of raw dataset by DCGAN which indirectly expressed by the distribution of DCGAN generated synthetic data; Step 2) Use the outputs from the convolutional layers in DCGAN discriminator on both raw dataset and synthetic dataset to represent their latent feature maps respectively, and use TDA on both the latent feature maps to visualize the difference in their distributions. In Step 2): The TDA Mapper generates a network representation of the feature space, in which each node corresponds to a set of data with similar features. The size of the node indicates the quantity of the data assigned in it while the color is used to distinguish whether the data is real or synthetic in the node. In ideal conditions, the synthetic data would have roughly equivalent proportion in each node. Accordingly, the categories of nodes are listed as follows:
    (a) No synthetic data are generated in this node; (b) both synthetic data and real data are placed in this node;
    (c) Synthetic data are generated anomalously outside of the real data feature distribution.
    (a) indicates the GAN (representation of DNN) is failing to generalize the data with the feature $\alpha$, $\beta$,
    (c) indicates that the GAN is still incomplete convergence, hence we will train the GAN further while (a) or (c) nodes occur.
    Part of (b) nodes containing few generated synthetic data with feature $\gamma$, thus we propose to emphasize our data collection procedure on data with $\gamma$.}
     \label{GAN-TDA framework (1,2)}
\end{figure*}

\paragraph{Topological data analysis}
For high-dimensional data analysis, dimension reduction approaches, such as PCA, MDS, t-SNE and $\etc$, are commonly applied \cite{DimensionReduction1,dimensionReduction2}. However, due to existing of generalization in NNs, the generated data may have more complicated topology than raw dataset in the high-dimensional feature space, while conventional methods fail to capture any structure from the data, which cause catastrophic lose in high-dimensional distance information for our analysis after dimension reduction. 
In our framework, topological data analysis (TDA) mapper is proposed to address this issue. With the key idea of multidimensional persistence, TDA can capture data structure and then preserve the high-dimensional distance information with simplicial complex \cite{TDA,TDAnature,TDAmapper}.

In our experiment, TDA is applied to tell the difference between the potential FD of data from raw dataset and that of synthetic data from the generator in GAN. With the high-dimensional clustering in TDA, discrete nodes are used to represent the original continuous feature distribution, while the connection between nodes indicates the distance in feature space. Before we do TDA, we mix these two dataset into one with data labels attached ($\ie$ real, synthetic), so that to maintain the consistency of the topological space in TDA. Then, by analyzing the proportion of real/synthetic data in each node, one can discover the weak nodes, which lack the synthetic data, and then identify the required data by the real data tags in these nodes to guide the new data collection.

It is worth noting that, in our proposed methodology, the description for required training data cannot exceed human knowledge about the data. Although GAN-TDA approach works on the feature space in NNs, the result is still interpreted using the human feature space (tags), which may not match the need of neurons. What we can do here is to tag data with our best knowledge, so that to make the data collection guidance more targeted.

\paragraph{Contribution and novelty}

In this paper, we propose GAN-TDA to identify which specific new data would be important for discriminative NNs' performance improvement based on an existing drone dataset, as the guidance to the new data collection work.  To our best knowledge, this paper is the first to reveal the relationship between NN's generalization ability and the composition of training data, so that to improve the model performance via newly collected data.

We make three major contributions:

(i) GAN-TDA framework is proposed to guide the new data collection. Specifically, we use GAN to capture the feature distribution in inference generalized by discriminative NNs from the training data, and TDA to identify the generalization weakness on the training data.

(ii) A drone image dataset using both real drone images as well as simulated images in CAD is established for our experiment. Each image is tagged with the drone's features ($\eg$ model, color, frame shape, camera position...) as many as we can.

(iii) We demonstrate our results on a drone image dataset, which contains both real drone images as well as simulated images from computer-aided design. When compared to random data collection (usual practice - discriminator accuracy of 94.67\% after 200 epochs), our proposed GAN-TDA informed data collection method offers a significant 4\% improvement (99.42\% after 200 epochs).

The remainder of this paper is organised as follows. In Section II, we demonstrate the working flow of our proposed GAN-TDA framework. In Section III, we apply our model on drone picture data for evaluation and validation. Section IV concludes this paper and proposes the ideas for future work.

\section{Method}

Given a dataset with clear properties labelled in tags that are detailed enough to guide the direction of new data collectivity (\eg images in a vehicle dataset labelled with the vehicle's maker, model, type, color, number of wheel \etc), our methodology is to identify which kind of data the discriminative model has weak generalization on by viewing the data distribution in the dataset and distribution learned by deep learning (DL) models.

\subsection{Step 1) - Learn data distribution by GAN}
The first step is to find the high-dimensional distribution of the raw dataset (drone images). According to the demonstration of $Step\ 1)$ in Fig. \ref{GAN-TDA framework (1,2)}, two networks named Generator $G$ and Discriminator $D$ with different network structures will be established and initialized with individual aim to generate synthetic images and to recognise the input image is real or synthetic. Before the training of GAN, all the image from the raw dataset will firstly be pre-processed into the same size (\eg 64*64 pixels) and secondly be normalized into the same scale. These steps are to ensure each input parameter (pixel intensities in each color channel) has a similar data distribution to guarantee the convergence of DL models\cite{BN}. 

Processed images from the raw dataset (real image dataset) will be divided into batches $B = \{B_1, B_2 ... B_i\} $ with a fixed size (\eg 64 pictures per batch). 
During the training, a batch of initialized noise set $n$ that follows a certain distribution (\eg Gaussian Noise) will be generated whose batch size is the same as data batch size (\eg 64*[100 samples from Gaussian Noise]). The initialized noise set will be re-produced at the beginning of each training iteration and then be processed into a set of synthetic images $G(n)$ by fractionally-strided convolutions \cite{DCGAN} in $G$.
The optimization of $G$ is expressed by minimizing the generative loss, which could be quantified by the discriminative result from $D$. And the optimization of $D$ is to minimize the discriminative loss on both synthetic images and raw images.

During the quantification of model loss, raw images $B_i$ will be labeled as $True$ and generated synthetic data $G(n)$ will be labeled as $False$, these information are stored into the real and synthetic tags matrix $T_{tag}(G(n),B_i)$ (mixed dataset). 
$D$ will scoring the real and synthetic rate on both $B_i$ and $G(n)$, the generated scores are stored in the real or synthetic score matrix as $T_{prediction}(G(n),B_i) = [D(G(n)),D(B_i)]$. 
The generative and discriminative loss could be further expressed by the divergence between $T_{tag}(G(n))$ and $T_{prediction}(G(n))$ and that between $T_{tag}(G(n),B_i)$ and $T_{prediction}(G(n),B_i)$. The divergence quantification uses Wasserstein distance to guarantee the stability of model training and avoid the collapse mode issue\cite{WGAN}.

The processes of training a DCGAN with Wasserstein distance is shown as Algorithm \ref{ALG_GAN}. 
During the model training, optimizing model parameters by backpropagation in both $G$ and $D$ will let the distribution of generated synthetic data $G(n)$ gradually approaching that of the real dataset $B$.
After the model converges, the distribution of the raw dataset is considered to have been learned and captured by GAN that the distributions of the GAN generated synthetic image set and the raw dataset is in a high similarity. Simultaneously, the ability to convert the given certain distribution noise set into real-enough synthetic data which follows the distribution of the raw dataset will be hiddenly stored with the form of model parameters in the GAN generator.

\begin{algorithm} 
	\caption{Traing of DCGAN with Wasserstein distance} 
	\label{ALG_GAN} 
	\begin{algorithmic}
	    \REQUIRE Training set $B = \{B_1, B_2 ... B_i\}$
		\REQUIRE Generator $G$, Discriminator $D$, Gaussian Noise $n$,
		    Wasserstein distance calculator $W$, Optimizer $Opt$
		\REQUIRE Number of total epochs $j$
		\STATE Initialize $G$, $D$
		\FOR{epoch $<= j$}
		    \FOR{$B_i \in  B$}
		    \STATE Initialize Gaussian Noise $n$
		    \STATE $T_{tag}(G(n),B_i) = [False*G(n), True*B_i]$
		    \STATE $T_{tag}(G(n)) = [False*G(n)]$
		    \STATE $T_{prediction}(G(n),B_i) = [D(G(n)),D(B_i)]$
		    \STATE $T_{prediction}(G(n)) = [D(G(n))]$
		    \STATE $loss\_D = W(T_{tag}(G(n),B_i),T_{prediction}(G(n),B_i))$
		    \STATE $loss\_G = W(T_{tag}(G(n)),T_{prediction}(G(n)))$
		    \STATE $D \gets Opt(loss\_D, D)$
		    \STATE $G \gets Opt(loss\_G, G)$
		\ENDFOR
		\STATE epoch = epoch + 1
    \ENDFOR
    \RETURN $D$, $G$
	\end{algorithmic} 
\end{algorithm}

\subsection{Step 2) - Latent feature maps extraction and TDA}

In CNN, convolutional kernels are designed with weight sharing property, and each multidimensional kernel representing a unique hidden feature. So, in the convolution process, the output of each convolutional kernel means the degree of activation of an image property in a series of adjacent receptive fields controlled by stride.

The latent feature map of an image contains a set of activation degrees on different high-dimensional features and each value in the latent feature map can be seen as the activation intensity of different convolution kernels. It can also be regarded as the expression of high-dimensional image features to a low-dimensional space like the encoded latent features by generative models like VAE. 
The latent feature map can express the characteristics of the image to a certain extent, and be used to reduce the image dimension to facilitate the analysis of the image. 

Certain data distribution will be expressed differently by latent feature spaces formed from different convolutional layers. To appropriately express the feature space for the raw dataset distribution, a proper multi-dimensional scale needs to be confirmed with a set of latent features learned by convolutional kernels. According to the fact that front convolutional layers tend to learn lower-dimensional visual features (\eg{line, Polyline, arc}), later convolutional layers trying to extract high-dimensional latent features which contain complex information about positioning and relationship based on features extracted by the front layers. The feature space formed by kernels from the last convolutional layer is chosen to keep the deeply high-dimensional global information used for feedforward layers inferences. And this space could clearly express the distance between distributions of the synthetic dataset and raw dataset.

As shown in Algorithm \ref{ALG_latent_feature_maps}, the latent feature map of a given image is defined as the output of GAN-discriminator's last convolutional layer after feeding the image into $D$ (Step 2 of Fig.\ref{GAN-TDA framework (1,2)}). Suppose the convolutional layers in $D$ is defined as a layer set $C = \{c_1, c_2 ... c_k\}$ and the generated synthetic image from $G$ as $S = \{G(n_1), G(n_2)...G(n_i)\}$ ($i$ is the batch number in $B$: to guarantee the generated synthetic images have the same amount of data as raw dataset). The tags of all synthetic image is set to $False$ as $T_{tags}(S) = [S*False]$, and $True$ for all images from real dataset $B$ as $T_{tags}(B) = [B*True]$. As shown in Algorithm \ref{ALG_latent_feature_maps}, the algorithm will return a dataset that contains both the latent feature maps for synthetic images $L_S$ and real images $L_B$ with their tags $T_{tags}(S)$ and $T_{tags}(B)$ for further TDA use.

\begin{algorithm} 
	\caption{Extraction of latent feature maps} 
	\label{ALG_latent_feature_maps} 
	\begin{algorithmic}
	    \REQUIRE Real image dataset $B = \{B_1, B_2 ... B_i\}$, Synthetic image dataset $S = \{G(n_1), G(n_2)...G(n_i)\}$
		\REQUIRE Convolutional layers of DCGAN's discriminator $C = \{c_1, c_2 ... c_k\}$
		\REQUIRE Tags for synthetic images $T_{tags}(S)$, and for real images $T_{tags}(B)$
		\STATE Define $L_S, L_B = \{\},\{\}$
		\FOR{$i <= len(B)$}
		    \STATE $l_B = B_i$
		    \STATE $l_{S} = G(n_i)$
		    \FOR{$j <= len(C)$}
		    \STATE $l_{B} = c_j(l_B)$
		    \STATE $l_{S} = c_j(l_{S})$
		    \STATE $j = j+1$
		    \ENDFOR
		\STATE$L_S.append(l_{S})$
		\STATE$L_B.append(l_{B})$
		\STATE $i = i+1$
        \ENDFOR
    \RETURN $\{L_S, L_B, T_{tags}(S), T_{tags}(B)\}$
	\end{algorithmic} 
\end{algorithm}

The viewing of data distribution in the chosen feature space is barricaded by its high dimensionality, where TDA makes reasonable dimensionality reduction representation to help the discovery of distance between two high-dimensional data distributions. We use Kepler Mapper for TDA visualization \cite{TDAmapper}.

\subsection{TDA interpretation}
In perfect training, the percentage of synthetic data in each node should be close, which means discriminative NNs give roughly equivalent generalization to every models. In practical, there always has bias on the generalization to different models, which can be observed from uneven percentage in each node. Low percentage indicates the lack of generalization and vice versa.

\subsection{Experimental setup}
Recent deep learning drone detection models make inferences mainly based on images and video information captured by surrounding cameras. These captured visual data will be processed to complex high-dimensional latent features by convolutional layers based on image properties of different receptive fields for further inference steps completed by feed-forward layers. Therefore, various appearance and shape factors of drones designed according to different working environments and purposes challenge CNN based drone identification models in recognition precision, generalization and robustness a lot.


To investigate which design property of drone is hard for general CNN-based classifiers to learn and discriminate, an experiment is established to apply the GAN-TDA method into a collected drone image dataset (raw image dataset). This experiment aims to prove that the drone discriminator trained using additional images collected with the GAN-TDA method performs better than the model using new images collected with the randomly method. During the experiment, the GAN-TDA model will analyze the raw dataset and generate the guidance on which kind of data should be collected additionally. Two models with the same net settings will be trained using different newly collected datasets (see \textit{Method 1)} and \textit{Method 2)} in Fig.\ref{intro}.\textit{III}) in a control experiment to evaluate the feasibility of GAN-TDA guidance. The datasets for two groups are: 

\textit{Group GAN-TDA - data from the raw dataset and additional data collected under guidance from GAN-TDA (additional data for several drone models)};

\textit{Group Random - data from the raw dataset and additional data collected averagely for all data categories (additional data for all drone models).}

During the validation, two models trained by Group GAN-TDA dataset and Group Random dataset respectively will be tested on the same validation dataset which contains images for all drone models distinct from the images in training sets (Group GAN-TDA and Group Random dataset). According to the performance comparison between the model trained with Group GAN-TDA dataset and with Group Random dataset, the superiority of GAN-TDA guided data in improving general target CNN-based model generalization could be viewed. Details are listed as follows. 

\begin{figure*}[!htbp]
     \centering
     \includegraphics[width=1.0\linewidth]{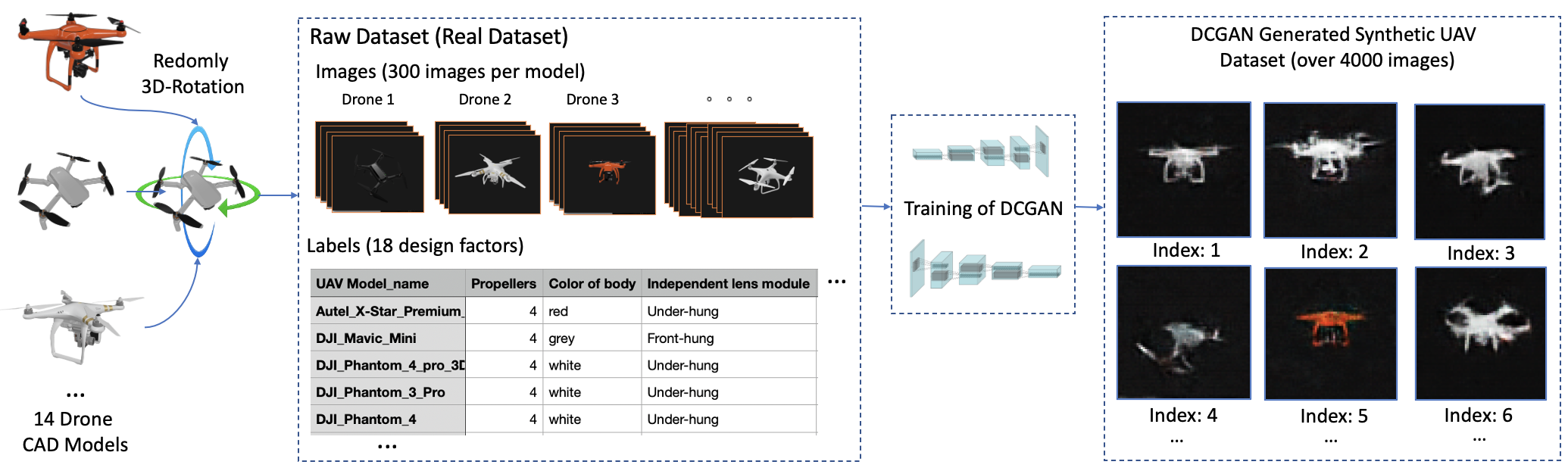}
     \caption{The Generating of the Raw Dataset and Synthetic Dataset}
     \label{Data}
\end{figure*}

\subsection{The dataset and hardware}

The raw dataset contains over 4000 pictures averagely collected from 14 popular commercial drones' 3D models (e.g., DJI Phantom 3, Phantom 4 and Phantom 4 pro). As shown in Fig.\ref{Data}, to simulate the real pictures of flying drones caught by monitors and cameras with different angles, we randomly rotate these 3D models on $x,y,z$ axis and take screenshots with drone centre-placed and 1800*1500 pixels resolution. During the data collection, the background of each drone model is set into black without ambient light effect to remove the high-frequency information from the background. These images are store in different folders named by their drone model's name, with an additional label file that contains a unique image ID for each collected drone image with 18 different hardware and appearance characteristics of these drones (e.g., shape of propellers, number of propellers, position of floor stand).

The experiment environment is split into two part: the training of DCGAN is transferred into Cloud served by Google Cloud Platform with a VM (Ubuntu 16.04) established and 1 Tesla V100 GPU (NVIDIA-SMI 450.102.04, CUDA 11.0, 16G Memory) embedded; the evaluation experiment is processed by 8-Core Intel Core i9 (16G Memory).

\subsection{DCGAN settings}

To accelerate the training speed of GAN and avoid the missing of meaningful details in the appearance, each image in the real dataset is resized into a resolution of 3*64*64 (R, G, B channels, 64*64 pixels) and processed with pixel value normalization in each channel (mean = 0.5, standard deviation = 0.5). As shown in Table \ref{DCGAN net}, the DCGAN-generator is designed with five fractionally-strided convolution layers to generate synthetic images with the same resolution as the pre-processed real image (3*64*64). The discriminator is designed with five convolutional layers which accept 3*64*64 images as input, and the activation function of the last layer (Sigmoid in original DCGAN \cite{DCGAN}) is removed to meet the requirements of use Wasserstein Loss \cite{WGAN}. During the training of DCGAN on the raw dataset, images from the dataset will be split into batches with 64 images each and shuffled before each training epoch. Gaussian noise is chosen to provide GAN-generator with original input, and in each batch training, a randomly initialized 64*100 noise set will be processed in GAN-generator into 64 synthetic images. According to the indications from \cite{GAN_training}, discriminator should be trained before generator and with more epochs than generator. The training rate is set to five that generator will be trained once after five times training of discriminator \cite{WGAN} with the same training rate $3e^{-4}$ using Adam Optimizer. The generating quality of generator will be checked every 400 epochs training, and part of the synthetic images are sampled and shown in Fig.\ref{Data} to demonstrate the generating quality of generator after 5000 epochs training.

\begin{table*}[!htbp]

\centering
\setlength{\abovecaptionskip}{0pt}%
\setlength{\belowcaptionskip}{10pt}%
\begin{tabular}{|c|c|c|c|c|c|}
\hline
\multicolumn{3}{|c|}{Generator}& \multicolumn{3}{c|}{Discriminator}\\

\hline
Layer&Type&Size&Layer&Type&Size\\
\hline
Input& Gussian Noise&100&Input& Image& 3*64*64\\
\hline
ConvTranspose 1& 4*4 Fractionally-strided Convolutions&512&Convolution 1& 4*4 Convolutions& 32\\
\hline
-& Batch Normalization & 512&-&Leaky ReLU&32*32*32\\
-& ReLU & 512*4*4&-&-&-\\
\hline
ConvTranspose 2& 4*4 Fractionally-strided Convolutions&256&Convolution 2& 4*4 Convolutions& 64\\
\hline
-& Batch Normalization & 256&-& Batch Normalization & 64\\
-& ReLU & 256*8*8& -&Leaky ReLU & 64*16*16\\
\hline
ConvTranspose 3& 4*4 Fractionally-strided Convolutions&128&Convolution 3& 4*4 Convolutions& 128\\
\hline
-& Batch Normalization & 128&-& Batch Normalization & 128\\
-& ReLU & 128*16*16& -&Leaky ReLU & 128*8*8\\
\hline
ConvTranspose 4& 4*4 Fractionally-strided Convolutions&64&Convolution 4& 4*4 Convolutions& 256\\
\hline
-& Batch Normalization & 64&-& Batch Normalization & 256\\
-& ReLU & 64*32*32& -&Leaky ReLU & 256*4*4\\
\hline
ConvTranspose 5& 4*4 Fractionally-strided Convolutions&3&Convolution 5& 4*4 Convolutions& 1\\
\hline
Output & Tanh&3*64*64&Output&None Activation Function/Sigmoid (validation)&1\\
\hline
\end{tabular}
\caption{DCGAN network training/validation settings}
\label{DCGAN net}
\end{table*}

\subsection{Validation settings}
The evaluation of the method is to show the performances of models trained with different additional datasets on distinguishing the synthetic and real drone data.

To control the experiment variable (avoid influences brought by different initialization ways), the network in DCGAN-discriminator is chosen to be the identical network initialization whose convolutional layers are pre-trained on the raw dataset. During the training of DCGAN, the increasing of discriminative ability in discriminator is suppressed by the gradually increasing adversarial power from DCGAN-generator. Once the generator is fixed, the training of the discriminator will no longer be limited and be seen as the training of a general discriminative DNN model. 

Based on the result from GAN-TDA (details are listed in the Section: TDA Result), we collect two additional datasets for Group GAN-TDA and Group Random respectively. The additional data for Group GAN-TDA is evenly collected from the 3 detected hard-to-learn drone models (DJI Phantom 3 Pro, DJI Phantom 4 and DJI Phantom 4 pro) and Group Random additional data is evenly collected from all drone models. In the practice environment, data accessibility for different models varies. Therefore, the evenly collected method is used in this experiment to represent the general randomly collection of new data.
To control variables, we collect 140 additional images per hard-to-learned drone model for Group GAN-TDA and 30 additional images among all 14 drone models for Group Random (the newly collected data accounts for 10\% of the raw dataset). With the use of discriminator as the initialization, the network structures of both models trained by Group GAN-TDA dataset and Group Random dataset are the same which shown in Table \ref{DCGAN net}, but Sigmoid activation function is added to the last feed-forward layer for the binary classification task.

The validation set is collected on all drone models evenly (50 images per drone model) and independent from the raw dataset and any additional dataset. Each image in the validation set will be pre-processed as the training sets before model inference. During training, model performance on the validation set will be supervised in each epoch.

\section{Results}

\begin{figure*}
     \centering
     \includegraphics[width=1.0\linewidth]{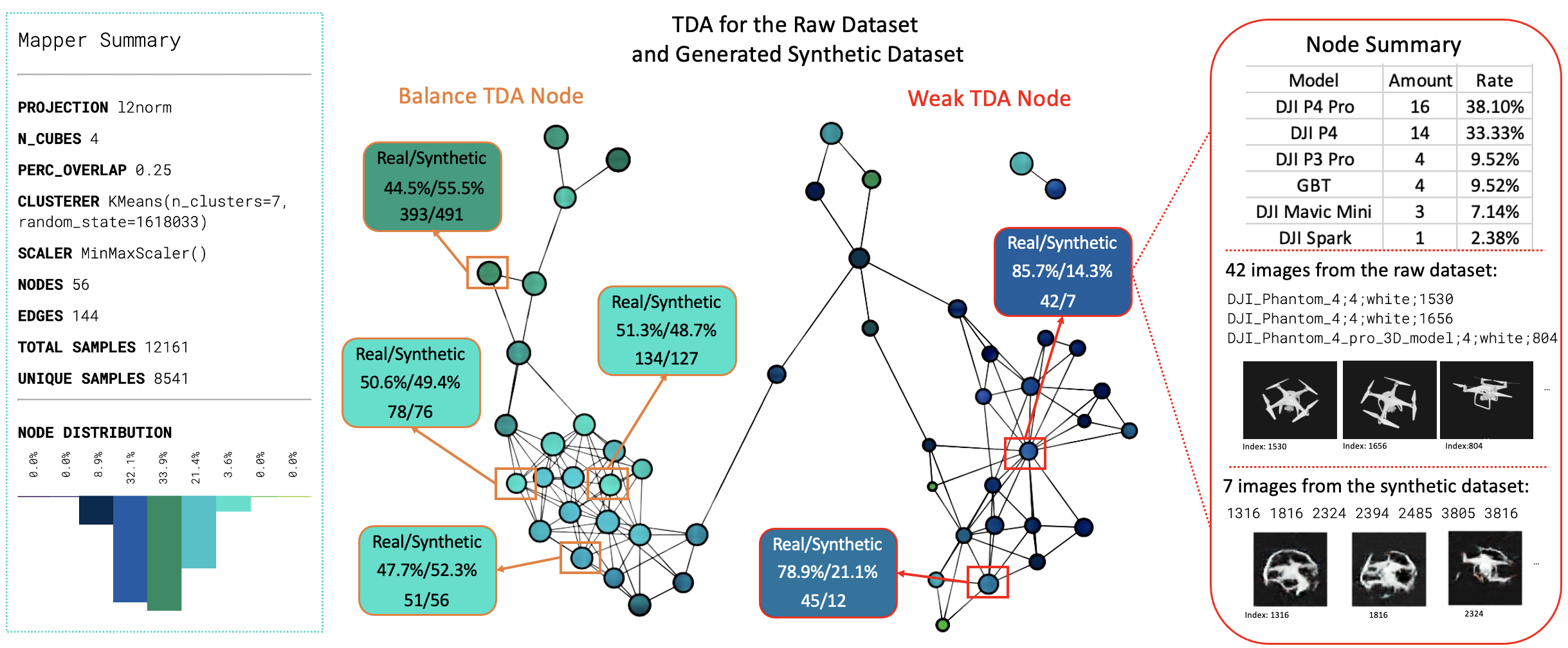}
     \caption{TDA output and analysis}
     \label{TDA_result}
\end{figure*}

\subsection{TDA result}
The outcome by TDA shows a topological analysis result to the distribution of both synthetic data and real data. By analyzing the TDA diagram shown in Fig.\ref{TDA_result}, we found three drone models out of 14 are more difficult for deep learning models to learn.

According to the \textit{mapper summary}, we can see none (a) or (c) type TDA node (clarified in Fig.\ref{GAN-TDA framework (1,2)}) occurs that the trained GAN do capture the data distribution of the raw dataset. There are two kinds of type (b) TDA nodes: {Balance TDA node (the amount of real data and synthetic data is balance) and Weak TDA node (few synthetic data in this node compared with real data)}.
The color of nodes reflects the internal balance of data that dark color refers to weak TDA nodes with low internal data balance. The rate of real data and synthetic data in some selected TDA nodes are listed in Fig.\ref{TDA_result} for demonstration.

According to the node distribution from the mapper summary, 8.9\% TDA nodes are in dark blue color so we will focus more on these weak nodes. Tracing the origin data whose latent feature map is placed in weak TDA points (which drone model are these data from) could provide clear guidance for new data collecting. We also list the details of an example weak TDA node as shown in Fig.\ref{TDA_result} - \textit{Node Summary}. From here, the proportion of real data from different drone models will be summarized (\eg{16 out of 42 real data are from drone model DJI Phantom 4 Pro, and count 38.1\% of real data in this node}).

By analyzing the data in all weak TDA nodes (with dark blue color), we found that the majority of real data placed in these nodes come from 3 drone models - DJI Phantom 3 Pro (22.7\%), DJI Phantom 4 (27.6\%) and DJI Phantom 4 (33.5\%). The result shows the DCGAN generalization ability on these drone models is weak. The TDA result further forms the GAN-TDA guidance that new data collection should focus on these models.

\subsection{Discriminator result}
The discriminator result shows that, on the designed validation dataset, the drone detection ability of the model trained with GAN-TDA guided additional data is better than the model trained by randomly collected additional data.
\begin{figure}
     \centering
     \includegraphics[width=1.0\linewidth]{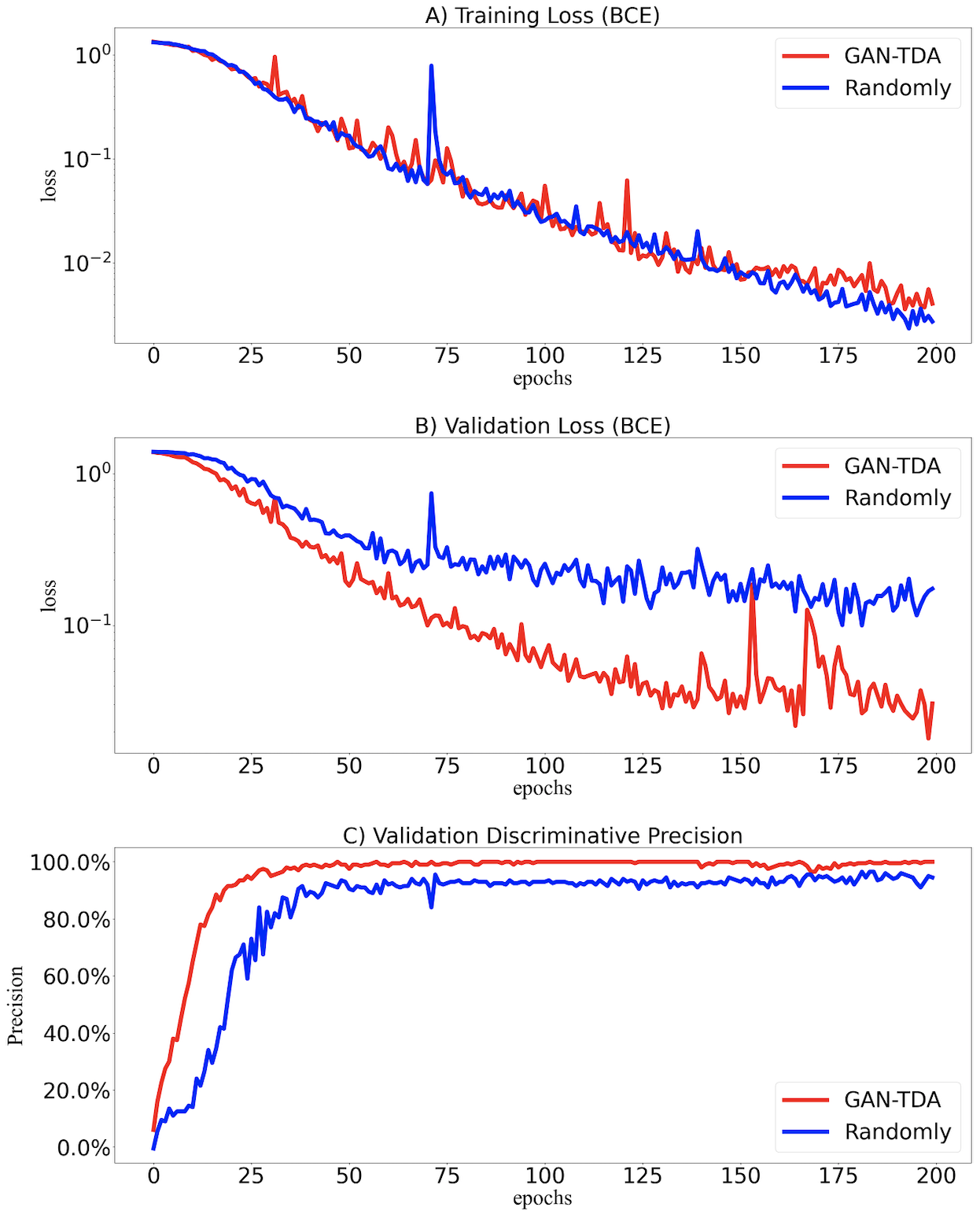}
     \caption{The performance comparison of targeted data collection method and random data collection method}
     \label{verification}
\end{figure}

The demonstration of the discriminator result is made in three parts: \textit{A) Training loss in binary cross-entropy (BCE); B) Validation Loss in BCE; C) Discriminative precision on the validation dataset}. 

According to \textit{A)} in Fig.\ref{verification}, the additional data collected by different collection methods do not generate obviously influences on model training loss. This means, once the network settings and additional data amount are fixed, the learning speed of the deep learning model on instances from the training dataset will not be influenced. 

But, as the models' BCE losses on the validation dataset are shown in \textit{B)}, the performance of model trained by Group GAN-TDA dataset shows a significant advantage than that of model trained by Group Random dataset in BCE loss (Final loss: 0.0305 - 0.1740). This means the models' generalization ability on unseen new data could be affected by the quality of additional data for training. In other words, although both collection methods guarantee the learning of instances from the raw dataset, GAN-TDA guided additional data performs better in boosting the model's learning of data distribution, which leads to the reinforcing of model generalization ability.

The result in \textit{C)} demonstrates the models' discriminative precision on validation dataset. The model trained with GAN-TDA guided additional data shows a quicker rising in inference precision of validation dataset compared with randomly collected data. And by the end of our training, GAN-TDA guided model's final precision on the validation dataset achieve a 4.75\% increase on that of model trained by random additional data (99.42\% - 94.67\% on 350 images). Simultaneously, GAN-TDA guided additional data could make the deep learning model achieve high accuracy faster than randomly collected additional data.

\section{Conclusion}

Increased proliferation of drones and autonomous air vehicles can disrupt critical national services (e.g., Gatwick Airport 2018). The economic damage for air transport is estimated to be millions per day for airports and airlines. Protection against drones is critical to ensuring smooth operation of services, whilst safeguarding it against the most severe threats. High resolution cameras using deep learning is challenged by the lack of training data sets. This often means a large amount of resources and time is dedicated to broad data collection and (re)training the neural network - without a guaranteed convergence in improving accuracy. This paper has used explainable deep learning to identify the missing data and guide data collection and generation. 

For high-dimensional image data, the GAN-TDA provide a solution to extract the latent features of each data instance as feature maps and generate a demonstration of the generalization ability of the convolution kernels on different latent features. With the mapping relationship among images, latent features and labels, the generalization ability of kernels on latent features could indicate that on different image properties (according to data labels). During model training, the training of hard-to-learned kernels with slow improvement in generalization abilities needs more training epochs and additional date feedings, which means an image instance with properties of these hard-to-learned kernels are more difficult for DL models to learn (in our experiment: images with these properties are hard for GAN to generate). Afterwards, analysing these learning-hardly images with their tagged properties can indicate the direction of new data collection. By applying GAN-TDA proposed in our paper, we achieve a 4.75\% precision boosting (99.42\%) on drone discriminative NN compared with a control Model which uses randomly collection method (94.67\%). Simultaneously, GAN-TDA guided data make the discriminative NN achieve the same inference performance with less training time.

\bibliographystyle{IEEEtran}
	
\bibliography{references}

\end{document}